\documentclass{article}
\pdfoutput=1

\usepackage{arxiv}
\usepackage[utf8]{inputenc} 
\usepackage[T1]{fontenc}    
\usepackage{hyperref}       
\usepackage{url}            
\usepackage{booktabs}       
\usepackage{amsfonts}       
\usepackage{nicefrac}       
\usepackage{microtype}      
\usepackage{lipsum}
\usepackage{graphicx}
\graphicspath{ {./images/} }
\usepackage[dvipsnames]{xcolor}

\title{WHAT MAKES US CURIOUS? ANALYSIS OF A CORPUS OF OPEN-DOMAIN QUESTIONS}

\author{
 Zhaozhen Xu \\
  Intelligent Systems Laboratory\\
  University of Bristol\\
  Bristol, UK \\
  \texttt{zhaozhen.xu@bristol.ac.uk} \\
   \And
 Amelia Howarth \\
  We the Curious\\
  Bristol, UK \\
  \texttt{Amelia.Howarth@wethecurious.org} \\
  \And
 Nicole Briggs \\
  We the Curious\\
  Bristol, UK \\
  \texttt{Nicole.Briggs@wethecurious.org} \\
  \And
  Nello Cristianini \\
  Intelligent Systems Laboratory\\
  University of Bristol\\
  Bristol, UK \\
  \texttt{nello.cristianini@bristol.ac.uk} \\

}

\begin{document}
\maketitle
\begin{abstract}
Every day people ask short questions through smart devices or online forums to seek answers to all kinds of queries. With the increasing number of questions collected it becomes difficult to provide answers to each of them, which is one of the reasons behind the growing interest in automated question answering. Some questions are similar to existing ones that have already been answered, while others could be answered by an external knowledge source such as Wikipedia. An important question is what can be revealed by analysing a large set of questions. In 2017, “We the Curious” science centre in Bristol started a project to capture the curiosity of Bristolians: the project collected more than 10,000 questions on various topics. As no rules were given during collection, the questions are truly open-domain, and ranged across a variety of topics. One important aim for the science centre was to understand what concerns its visitors had beyond science, particularly on societal and cultural issues. We addressed this question by developing an Artificial Intelligence tool that can be used to perform various processing tasks: detection of equivalence between questions; detection of topic and type; and answering of the question. As we focused on the creation of a “generalist” tool, we trained it with labelled data from different datasets. We called the resulting model QBERT. This paper describes what information we extracted from the automated analysis of the WTC corpus of open-domain questions.
\end{abstract}


\section{Introduction}
In 2017 “Project What If” was started at the “We the Curious” (WTC) science-centre of Bristol (UK), with the stated intention of being the first exhibition all about “the curiosity of a city”. Its aim was no less than capturing the curiosity of Bristolians by collecting all their questions. It was focused on the questions “of real people”, and through these is aimed at understanding what Bristolians were curious about. In other words, it was not so much about the answers to individual questions, as it was about understanding a Community from the questions it asks.

Despite the clear identity of WTC as a science-centre, the organisers of this project were trying to gauge a broader set of concerns, about culture and society, in a time of rapid change. A collection of the spontaneous questions of thousands of people was expected to tell us a lot about the people who asked them.

Over the following three years, the project gathered over 10,000 questions, both in their “museum” venue and in initiatives around the city. That list taken together contained many questions, worries, doubts, and ambitions of thousands of citizens.

At the end, just one final question remained: what did that vast corpus contain? This is not something that can be answered by a single person reading the questions, but also not by a simple statistical analysis. What is needed is intelligent software capable of understanding the questions, their type and topic. As a further level of ambition, we asked: would the AI system be able to answer some of these questions?

We report here on the first content analysis of that set of questions, which was performed with Artificial Intelligence tools specifically created for that task. The AI algorithm was based on Deep Learning technologies and was tasked to solve three main problems: detecting topic of questions; detecting equivalent questions with similar meaning but different wording; locating potential answers to these same questions in Wikipedia.

The field of automated Question Answering (QA) is a new but fast-growing branch of AI, driven by commercial systems such as Alexa and Siri. According to a US report on smart speaker consumer adoption, 84.0\% of their users had tried to ask a question through the speaker, 66.0\% and 36.9\% did so on a monthly and daily basis respectively \cite{kinsella2019smart}. But at the core of any method for QA (as well as other question processing tasks) there is the challenge of representing a short sentence in a way that reflects its meaning. For this purpose, we made use of a deep-learning technique known as “BERT” \cite{devlin2018bert} which will be described below.

We discover that over half of these questions were about the topic of Science and Mathematics, and over a quarter were of the type WHY and HOW. These are known as factual questions and can sometimes be answered by automated systems, perhaps on the basis of Wikipedia. But what was more interesting was the large number of non-factual questions. For example, the counterfactual ones of the type IF which could not be handled in this way. 

The QA task itself can be described as an open-domain open-book question answering task, in that no limitation is posed a priori on the topic of the question, and the answering system is allowed to “look at the book” in order to answer. 

The main contributions of this article are: a general-purpose method to represent short questions, that is useful for a range of different tasks; and a statistical overview of the contents of the WTC corpus, enabled by that method. 

The article describes the WTC corpus in Section \ref{wtc}, the algorithm in Section \ref{QBERT}, the content-analysis of the corpus in Section 4, and the discussion of results in Section 5.

\section{WTC CORPUS OVERVIEW}
\label{wtc}
We will call our dataset of open-domain questions “the WTC corpus”, this section describes its origin and main features.

The dataset is originated from a project run in Bristol (UK) by “We The Curious” (henceforth WTC), an educational charity and science centre.

Between January 2017 and October 2019, WTC collected over 10,000 open-domain questions from a diversity of sources: onsite (at the WTC venue in Bristol), offsite, and online. Offsite question gathering ensured questions were received from various Bristol postcodes (BS1 – BS16), and general submissions were received from all remaining postcodes. These refer to questions collected in the venue, written on cards by visitors, later stored and entered manually into the question database. 

Based on this initiative, WTC created a digital database of questions, which is in WTC’s possession. WTC is responsible and accountable for protecting the personal data of individuals submitting this information alongside their questions. All personal data is held by WTC in compliance with GDPR protocol and personal data is not shared with other parties, including the analysis team of this project. For the purpose of the present study a smaller dataset was generated, by removing all the personal data that was associated to the questions, and only this was shared with the analysts (ZX and NC). 

Manual Curation of the WTC Corpus. The raw corpus also included repeated questions, various types and topics, and other non-question sentences. Questions were first moderated manually by WTC staff. The questions in the database were also screened for any possible identifying information or potentially offensive or inappropriate language or content. These were removed from the database. After moderation, the resulting dataset contained 10,073 questions.

This second, anonymised and moderated, textual dataset is what we will call the WTC-corpus in this paper. 

\textbf{Automated Pre-processing:} Some simple pre-processing was performed before content analysis, such as removing exactly identical questions and questions shorter than three words. After these steps, the filtered WTC dataset contained 8.600 questions, using 5,732 words. The length of questions is between 3 and 55 words, with an average of 7.15 words. 87.96\% of the questions are within 10 words.

\begin{figure}[hbt] 
    \centering
    \includegraphics{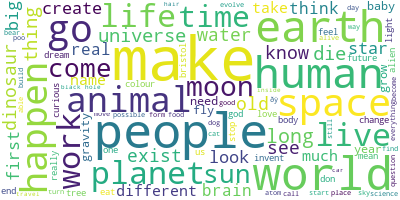}
    \caption{The word cloud is generated from the curated and filtered WTC corpus. The words were lemmatised before generating the graph. The size of the word is proportional to its frequency in the corpus.}
    \label{fig:word}
\end{figure}

The word cloud in figure \ref{fig:word} shows that the questions cover universe and space, human body, energy and climate change, animals and plants, chemistry and materials, the future and some other topics outside of the typical science categories listed before. We also identified 5,022 “equivalent” question pairs, as will be described in section 4.2.

\section{QBERT: A MULTI-TASK QUESTION-PROCESSING VERSION OF BERT}
\label{QBERT}
In order to process the questions in our corpus we embedded them into a vector space of 768 dimensions, using a model based on the BERT method.

In particular, we fine-tuned a BERT based model (sentence-BERT, or S-BERT \cite{reimers2019sentence}) by using a diverse set of question-related datasets, which will be described below. For convenience in our experimental comparisons, we named this refined BERT model “QBERT” to indicate that it was specifically fine-tuned to handle questions. 

BERT is a standard method for the representation of sentences, based on the technology of Transformers (more specifically, multiple stacked encoders) as described in \cite{devlin2018bert}. It adds a classification token [CLS] at the beginning of the input sequence and encodes the input sequence by assembling its token embeddings, segment embeddings, and position embeddings. The bidirectional transformer encoder \cite{vaswani2017attention} is then trained with two unsupervised tasks: masked language model and next sentence prediction. The encoder’s output can be used for downstream tasks like classification, question answering, and sentence tagging.

Fine-tuning is an important step in using BERT, as it adapts the model to the specific class of sub-tasks at hand. To train a model which can understand the content of the corpus and find possible answers to the questions, we used three kinds of tasks to fine-tune the standard BERT model: Question-Equivalence (QE), Question-Answering (QA), and Question-Topic (QT), defined as follows.

\begin{itemize}
    \item QE BERT is given two questions and is required to decide whether they are equivalent.
    \item QA BERT is given a question and a set of candidate answers and is required to decide which of them is the correct answer.
    \item QT BERT is given a set of questions and topics and is required to determine the topic of each question.
\end{itemize}

We tuned the parameters of a pre-trained BERT on each of these tasks, using three different datasets that will be described below. We also measured its performance on each of these tasks separately. Our focus was not on achieving record performance on any of these tasks, but rather on creating a model that can perform well on each of them: rather than three specialists, we wanted a generalist model.

\subsection{The Method}
In order to train the model, we reduced the three NLP tasks described above to a series of standard classification tasks: QE determines if a pair of questions are equivalent or not; QA determines if a candidate answer is appropriate for a given question; and QT categorises the text by topic. 

\begin{figure}[hbt] 
    \centering
    \includegraphics[width=0.7\textwidth]{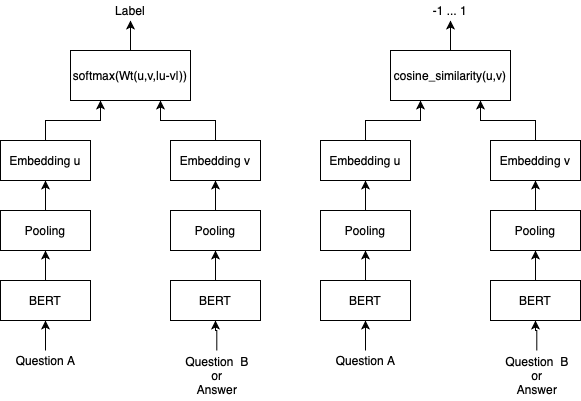}
    \caption{S-BERT architecture. Left: Training on classification task, Right: Inference by giving a cosine similarity between sentences. All the BERTs share the same parameters.}
    \label{fig:sbert}
\end{figure}

Recent success of pre-training language models proved that training and fine-tuning a single model could increase performance in different tasks \cite{devlin2018bert,radford2018improving,yang2019xlnet} Our approach is based on S-BERT \cite{reimers2019sentence}, a modified BERT that captures sentence similarity and provides an embedding for a given sentence. Comparing to the original BERT that uses the [CLS] token as sentence embedding, S-BERT applies a pooling method to compute the mean of all output vectors from BERT. In addition, S-BERT concatenates the sentence embeddings with the element-wise difference of the sentence pairs during training so that semantically similar sentences are close to each other in vector space. Another advantage of S-BERT is that it is more time-efficient in finding the most similar sentence while combining with Faiss \cite{JDH17}. As it is shown in figure \ref{fig:sbert}, while training on classification task, S-BERT calculates $Softmax(W_{t}\cdot (u, v, |u-v|))$ to predict the label for the sentence pairs. $W_{t}$ is a trainable weight, and $|u-v|$ is the element-wise difference of the embeddings. The BERTs in figure \ref{fig:sbert} share the same parameters during training. Through inference, S-BERT generates embeddings with the trained model. The distance between the embeddings can be measured with cosine distance.

\subsection{Training Datasets}
One of the main challenges for this research is the lack of labelled data. It is expensive to create reliable labels for each task. Thus, the model is trained and fine-tuned on some other existing datasets then transferred to analyse WTC. 

\textbf{Quora Question Pair (QQP) \cite{quora}} is a question pair identification competition first released on Quora in 2017. It contains 404,290 pairs of different questions from Quora with annotations. After embedded with S-BERT, there are 537,931 unique questions in the dataset. QQP competition intends to classify if the questions are duplicated, which is ideal for fine-tuning our model. By training on QQP, we figure out the distance threshold that can be transferred to analyse WTC corpus.

\textbf{WikiQA \cite{yang2015wikiqa}} is a question-answering dataset that extracts the questions from real-world query logs on Bing. All the questions in WikiQA are factual queries that start with Wh-word and have at least 5 users click on a Wikipedia page after searching. The answers are consist of candidate sentences from Wikipedia and human labelled as a correct answer or not. The dataset includes 3,047 questions and 26,154 sentences; 1,239 of the questions contain a correct answer from Wikipedia. Training with WikiQA enables us to evaluate our question answering system on open-domain.

\textbf{Yahoo! Answer \cite{zhang2015character}} is a corpus generated from Yahoo! Research Alliance Webscope program. The corpus contains 1,460,000 samples in 10 different topics. Each sample includes the topic, question title, question content, and the best answer. During training, Yahoo! Answer was separated into two datasets, Yahoo Topic (YT) and Yahoo Question-Answering (YQA). YT contains all the questions and categories used for QT training. YQA is made up of questions and the corresponding answers that are less than 35 words. There are 754,566 question-answer pairs in YQA.

\subsection{Training and Fine-tuning}
\label{model}
Following S-BERT, we trained the model with classification tasks. The basic S-BERT was only trained on natural language inference dataset and semantic textual similarity dataset that contain sentence pairs with labels such as SNLI \cite{bowman2015large}, NLI \cite{williams2017broad}, and STS \cite{cer2017semeval} dataset. Thus, it has poor performance in detecting similar question pairs. Following the architecture of S-BERT in figure \ref{fig:sbert}, we trained the sentence embedding model to learn the similarity between questions with data from QQP. The length was limited to 35 tokens for each input sequence because 99.93\% of the questions are shorter than 35 words in the WTC corpus. The sequences with more than 35 words were truncated after the limited length.

We have used the classification technique described in S-BERT \cite{reimers2019sentence} for QE and QA classification. For QQP and WikiQA datasets, the model took questions pairs or questions answer pairs as the input and produces the label in terms of 1 or 0. 1 represents that the input sequences are similar or related. Each input sequence was tokenised and embedded by BERT-base then produce an embedding with 768 dimensions. BERT-base used in this model is a smaller BERT version containing 12 layers and 110M parameters. All the weights in BERT were updated during training. Comparing to softmax loss in S-BERT, the contrastive loss is more capable of mapping the similar vector in high dimensional space into nearby points in a lower dimension \cite{hadsell2006dimensionality}. Hence, we minimised the online contrastive loss and optimised it by Adam optimiser with a learning rate of 2e-5. The contrastive loss combines loss from both positive samples and negative samples with a margin of 0.5. The margin ensures that negative samples have a more significant distance than the margin value. YQA was introduced as a supplement dataset for QA task because WikiQA did not have enough data considering the size of the model. Since YQA only contains corresponding question-answer pairs, multiple negatives ranking loss that requires only positive labels is applied instead of online contrastive loss. 

In QE and QA, instead of classifying if the sequences are related, it is more important that the system can retrieve all the related sequences for given questions. The problem is how to quantify `related’ with embeddings. A cosine similarity threshold was introduced in this model. First, all the sequences in the training set were embedded with the fine-tuned model. The sequence pairs were classified as positive if they have higher similarity than the threshold. We used 2 different strategies to decide the threshold for QE and QA. For identifying similar question pairs, the similarity threshold with the best accuracy in the QQP was found to quantify any questions pairs during training. On the other hand, for question-answer pairs, we leveraged the threshold with the best precision in WikiQA instead. While retrieving answers from the knowledge base, there are usually millions of candidates and we wanted the answer to be as reliable as possible. With both sequence embeddings and the threshold observed above, the model is capable to classify and search all the related sequences in the corpus by calculating the cosine similarity between sequences.

Contrary to QE and QA, QT took one question as input and predicted the question topic with the embedding. We have used the similar classification technique described in S-BERT \cite{reimers2019sentence}, but only one BERT was needed in the network. An additional softmax layer was applied after BERT to map the embedding into probability for each topic. We fine-tuned the trained BERT and the softmax layer with extra data in YT. 

QBERT was trained for 10 epochs, respectively for each dataset, with the training data divided by the data provider for each task. Except for YT, all the data was applied during training and trained for 5 epochs. For QE and QA, the network was trained with a batch size of 150. Moreover, for QT, the batch size was 350. The model was evaluated by accuracy for all classifications. In answer retrieval, we evaluated accuracy, precision and recall for the first answer.

In this experiment, the networks were trained with 1 GeForce GTX TITAN X GPU. It took 7 hours, 0.5 hours, 9 hours, 16.5 hours to train on QQP, WikiQA, YQA, and YT, respectively.

\subsection{Performance of QBERT}
The results for models fine-tuned with different datasets are illustrated in table \ref{tab:classification} and table \ref{tab:retrieval}. In QE and QA tasks, the embedding network was trained with pairs of sentences, so with more semantic textual similarity datasets, they both achieve better performance. However, for QT, it was trained with a different structure fine-tuned based on QE and QA results with the same YT dataset. QT task does not require the embedding model to capture the sentence similarity. Therefore, pre-training with more semantic textual similarity datasets does not significantly affect the result of QT. Furthermore, the fine-tuned QT model performs worse on other tasks because it only trained with one BERT and limited the performance on capture sentence similarity.

\begin{table}[hbt]
 \caption{Performance of QT, QE, and QA classification tasks. The models are trained with different datasets. The model’s name indicates the training sequence of each dataset.}
  \centering
\begin{tabular}{|c|c|c|c|c|c|c|c|c|}
\hline
                                                                   & \multicolumn{2}{c|}{\textbf{QT}}       & \multicolumn{3}{c|}{\textbf{QE}}                                              & \multicolumn{3}{c|}{\textbf{QA - Classification}}                                                      \\ \cline{2-9} 
                                                                   & \multicolumn{2}{c|}{\textbf{Accuracy}} &                                      & \multicolumn{2}{c|}{\textbf{Accuracy}} &                                      & \multicolumn{2}{c|}{\textbf{Accuracy}}                          \\ \cline{2-3} \cline{5-6} \cline{8-9} 
                                                 & \textbf{Train}     & \textbf{Test}     & {\textbf{Threshold}} & \textbf{Train}     & \textbf{Test}     & {\textbf{Threshold}} & \textbf{Train}                 & \textbf{Test}                  \\ \hline
\textbf{STS+NLI}                                                   & 85.71\%            & 72.44\%           & 0.800                                & 74.39\%            & 74.77\%           & {\color{red} 0.910}         & {\color{red} 94.89\%} & {\color{red} 95.24\%} \\ \hline
\textbf{QQP}                                                       & 82.86\%            & 72.32\%           & 0.850                                & 90.48\%            & 89.59\%           & -                                    & -                              & -                              \\ \hline
\textbf{QQP+WikiQA}                                                & 86.21\%            & 72.32\%           & 0.825                                & 91.42\%            & 85.44\%           & 0.713                                & {\underline{99.99\%}}                  & 94.70\%                        \\ \hline
\textbf{\begin{tabular}[c]{@{}c@{}}QQP+WikiQA+YQA\end{tabular}} & 86.57\%            & {\underline{72.51\%}}     & 0.825                                & 82.08\%            & 80.13\%           & 0.755                                & 95.16\%                        & 94.74\%                        \\ \hline
\textbf{YQA+WikiQA+QQP}                                            & {\underline{86.78\%}}      & 72.14\%           & {\underline{0.875}}                          & {\underline{99.20\%}}      & {\underline{90.29\%}}     & {\underline{0.797}}                          & 96.34\%                        & 93.92\%                        \\ \hline
\textbf{\begin{tabular}[c]{@{}c@{}}QQP+YQA+WikiQA\end{tabular}} & 86.58\%            & 72.37\%           & 0.850                                & 78.27\%            & 76.49\%           & 0.756                                & 99.97\%                        & {\underline{95.18\%}}                  \\ \hline
\end{tabular}
\label{tab:classification}
\end{table}

\begin{table}[hbt]
\caption{Performance of QA Retrieval tasks. The models are evaluated on the WIKIQA dataset. The model’s name indicates the training sequence of each dataset.}
  \centering
\begin{tabular}{|c|c|c|c|c|c|c|}
\hline
{\textbf{}}                                         & \multicolumn{6}{c|}{\textbf{QA - Retrieval}}                                                                                    \\ \cline{2-7} 
                                                                   & \textbf{Accuracy@1} & \textbf{Precision@1} & \textbf{Recall@1} & \textbf{Accuracy@1} & \textbf{Precision@1} & \textbf{Recall@1} \\ \cline{2-7} 
                                                                   & \multicolumn{3}{c|}{\textbf{Train}}                            & \multicolumn{3}{c|}{\textbf{Test}}                             \\ \hline
\textbf{STS+NLI}                                                   & 21.64\%             & 21.64\%              & 19.74\%           & 29.88\%             & 29.88\%              & 27.21\%           \\ \hline
\textbf{QQP}                                                       & -                   & -                    & -                 & -                   & -                    & -                 \\ \hline
\textbf{QQP+WikiQA}                                                & 82.06\%             & 82.06\%              & 78.53\%           & 28.63\%             & 28.63\%              & 27.07\%           \\ \hline
\textbf{QQP+WikiQA+YQA}                                            & 53.81\%             & 53.81\%              & 50.53\%           & {\underline{46.47\%}}       & {\underline{46.47\%}}        & {\underline{43.36\%}}     \\ \hline
\textbf{\begin{tabular}[c]{@{}c@{}}YQA+WikiQA+QQP\end{tabular}} & 60.92\%             & 60.92\%              & 57.48\%           & 37.76\%             & 37.76\%              & 34.85\%           \\ \hline
\textbf{\begin{tabular}[c]{@{}c@{}}QQP+YQA+WikiQA\end{tabular}} & {\underline{85.17\%}}       & {\underline{85.17\%}}        & {\underline{81.28\%}}     & 46.06\%             & 46.06\%              & 42.36\%           \\ \hline
\end{tabular}
\label{tab:retrieval}
\end{table}

While pre-training the sentence embedding model with all three datasets for QE, we observed that the order in which the training data are presented has a great effect on the final result. As shown in table \ref{tab:classification}, with the same datasets, the model trained with QQP as the last outperforms the model trained QQP first around 10.16\%, from 80.13\% to 90.29\%. Besides, the cosine similarity threshold increases from 0.825 to 0.875 means that the questions with similar meanings are closer to each other in vector space. 

Possible remedies to this effect will be the object of a separate study, as they are not relevant to the problem we are addressing in this paper.

In QA, we fine-tuned S-BERT on the classification task and evaluated it on both classification and retrieval tasks. Similar to QE, the best threshold contains more datasets from different tasks. In the classification task, the original S-BERT trained on STS+NLI outperforms other models. However, this is due to the bias of WikiQA dataset. 94.89\% of the question-answer pairs in the training set of WikiQA are labelled as 0, and 95.24\% in the test. S-BERT does not manage to identify the correct answer, and it only uses a large threshold to ensure that all the question-answer pairs are categorised as negative. WikiQA only contains question-answer pairs in the dataset. In order to evaluate the performance on retrieval task, a knowledge base that includes all the candidate sentences in WikiQA dataset was generated. And during evaluation, we leveraged only questions with a correct answer in the answer base. As shown in Table \ref{tab:retrieval}, in QA retrieval, training with extra YQA data dramatically increase the accuracy of the WikiQA test set.

According to QE and QA performance in Table 1 and Table 2, we noticed that the best models for each specific task might have poorer accuracy on another. Besides, both of the models with the best performance are over-fitted on the training set. Therefore, we used the model following the training sequence of QQP, WikiQA, YQA, which has a more balanced performance on all three tasks, as our multi-tasking generalist QBERT. We applied this generalist QBERT to our task of the content analysis of the WTC corpus.

\section{CONTENT ANALYSIS OF WTC CORPUS}
In the WTC corpus, we have questions with various content from different people. There are factual questions like “How are mirrors made?” and “Who built the internet/electricity?”; and counterfactual questions such as “How long will the earth and humans last if we carry on damaging it and nothing changes?” and “Would a car weigh more if there was a flying pigeon inside of it?”. They cover multiple topics and have overlap in the content. By using QBERT trained and fine-tuned with the public datasets YT, YQA, WIKIQA, and QQP, we also analysed our WTC corpus in the three different tasks of QT, QE, and QA.

\subsection{QT}
To classify the questions, we first identified each question’s type and topic. The type of the question was categorised into the following categories: WHAT, WHO, HOW, WHEN, WHERE, WHY, WHICH, and IF which will be further discussed below. For this first classification, we just used simple keyword-matching. On the other hand, the topic of the question was classified by the trained network in section \ref{model}. The topics included: Business \& Finance, Computers and Internet, Education \& Reference, Family \& Relationships, Health, Politics \& Government, Science \& Mathematics, Society \& Culture, Sports. 

It is important to notice that there are many non-scientific questions in this list, which was part of the initial intent of the overall project: to assess the scope and breadth of the curiosity of an entire community.

Notice also that we had 9 types and 10 topics, and therefore 90 Question “Themes” to which we could allocate the over-8000 distinct questions that have survived the various stages of filtering.

Besides the 7 WH-questions, and HOW, we have also defined a further class of questions that we call IF questions. The aim was to find a simple way to approximate the counterfactual questions of the type “what if”, which however are difficult to capture exactly by keyword matching, but can well be approximated in this context by checking for the use of the word “if”.

A counterfactual (CF) question is defined as a question of the type: “what would happen if X was true”. The understanding is that X is not a true fact, but the asker of the question is considering the possible consequences of X being true. This kind of question takes its name from being “counter to the facts”, is often used in defining the notion of Causality (e.g. in \cite{pearl2018causal}), and indicates a mental process directed at understanding the mechanism behind observations. 

Given a question, we assigned it to the type of the first keyword from our list that was found in it, with one notable exception described below. For example, the question “Why do we get butterflies when we like someone?” was categorised as WHY. However, questions that contain the keyword “if”, such as “what if” and “How ... if ...” were classified as IF questions. The category OTHER includes yes/no questions or sequences that do not fit into other categories.

Then we applied the topic classification model to identify 10 topics in WTC. Table \ref{contingency} shows the frequency distribution of the questions across types and topics. The most “asked” topics in the corpus are science \& mathematics and society \& culture, which make up 66.35\% of the corpus. Moreover, half of the questions are HOW and WHY questions.

\begin{table}[hbt]
\caption{Contingency table for topics the types in WTC.}
  \centering
\begin{tabular}{|l|l|l|l|l|l|l|l|l|l|l|}
\hline
                                 & \textbf{how} & \textbf{what} & \textbf{when} & \textbf{where} & \textbf{which} & \textbf{who} & \textbf{why} & \textbf{if} & \textbf{other} & \textbf{\%} \\ \hline
\textbf{Business \& Finance}     & 121          & 100           & 16            & 18             & 0              & 26           & 191          & 30          & 136            & 7.42                     \\ \hline
\textbf{Computers \& Internet}   & 34           & 9             & 3             & 2              & 0              & 3            & 18           & 5           & 34             & 1.26                     \\ \hline
\textbf{Education \& Reference}  & 132          & 81            & 8             & 11             & 2              & 50           & 84           & 16          & 68             & 5.26                     \\ \hline
\textbf{Entertainment \& Music} & 55           & 56            & 10            & 10             & 0              & 12           & 80           & 39          & 108            & 4.30                     \\ \hline
\textbf{Family \& Relationships} & 44           & 32            & 8             & 8              & 0              & 1            & 95           & 14          & 68             & 3.14                     \\ \hline
\textbf{Health}                  & 159          & 66            & 18            & 10             & 0              & 6            & 299          & 34          & 84             & 7.86                     \\ \hline
\textbf{Politics \& Government}  & 23           & 18            & 7             & 2              & 0              & 5            & 57           & 22          & 51             & 2.15                     \\ \hline
\textbf{Science \& Mathematics}  & 1355         & 646           & 88            & 99             & 15             & 58           & 1107         & 392         & 918            & 54.40                    \\ \hline
\textbf{Society \& Culture}      & 142          & 159           & 23            & 21             & 0              & 52           & 286          & 108         & 237            & 11.95                    \\ \hline
\textbf{Sports}                  & 47           & 14            & 5             & 0              & 0              & 15           & 48           & 7           & 59             & 2.27                     \\ \hline
\textbf{\%}          & 24.56        & 13.73         & 2.16          & 2.10           & 0.20           & 2.65         & 26.34        & 7.76        & 20.50          &                          \\ \hline
\end{tabular}
\label{contingency}
\end{table}

\subsection{QE}
Although the corpus was pre-processed to remove identical questions, there are still many “equivalent” questions left in the corpus. For the purpose of this study, we consider two questions as equivalent if they have the same answer. For instance, questions “What is our purpose?” and “What is the aim of our life?” have different wordings but they have the same answer. Finding equivalent questions in the corpus helps us to further understand any patterns, such as clusters, in the set.

To better understand the performance of finding equivalent questions in WTC, we randomly sampled 1,000 questions and generated a list of candidate question pairs. The questions were embedded by the S-BERT that trained only with the NLI dataset. The top 10 questions with the largest cosine similarity were selected as similar question pair candidates for each question. For duplicate question pairs such as [Q1, Q2] and [Q2, Q1], we only kept one of them for annotation. The question pairs were labelled with 0 or 1, where 0 represents different, and 1 for similar. The author labels 5,022 pairs of candidate questions, 728 pairs are similar, and 4,294 pairs are different. Table \ref{tab:wtc-example} provides some examples of the data we label.

\begin{table}[hbt]
\caption{Examples from labelled WTC question pairs.}
  \centering
\begin{tabular}{lllll}
\hline
\textbf{qid1} & \textbf{Question1}              & \textbf{qid2} & \textbf{Question2}           & \textbf{Label} \\ \hline
2             & Who is the richest?             & 6992          & Why can’t I be rich?         & 0              \\ \hline
33            & How do you make glass?          & 2001          & How is glass made            & 1              \\ \hline
50            & When did the humans come alive? & 7257          & When did humans first exist? & 1             
\end{tabular}
\label{tab:wtc-example}
\end{table}

When the cosine similarity threshold is 0.825, the model obtains 90.80\% accuracy on the sampling data. QBERT obtains a better accuracy on WTC with QQP, which has 80.13\% accuracy on the test set. This proves that QBERT can be applied to a corpus of unseen questions.

Moreover, we identified clusters by applying a “graph community detection” method \cite{clauset2004finding, hagberg2008exploring} in order to group similar questions. To cluster the questions, a graph is built with question nodes using the cosine distance matrix. An edge is added to the nodes if the distance between a pair of questions is smaller than $1-cosine\_similarity$. There are 6,060 communities found in the WTC corpus, which represents 6,060 different questions in the corpus. Of these, 5,398  questions do not have any similar question in the corpus.

\subsection{QA}
We are also interested in whether WTC questions can be answered (with high confidence) by the QBERT model. The model aims to retrieve a sentence as the answer from an unstructured knowledge base.

For WTC, we used Wikipedia summary \cite{scheepers2017compositionality} as a knowledge source during inference. The corpus includes the title and the first paragraph as the summary for each Wikipedia article extracted in September 2017. The raw texts of the Wikipedia have 116M sentences initially. Of these, 22M are in the summaries. After we embedded with QBERT, 21M sentences have different embeddings. The summary of Wikipedia provides the article’s primary information. In the meanwhile, the summary reduces about 80\% of the sentences from the original Wikipedia. 

In order to retrieve the answer from Wikipedia, we calculated the average distance of correct answers in the QA retrieval task described in Section 3.4. The best-scored sentence from Wikipedia Summary with a higher similarity than 0.688 was considered as the answer for given question.

In order to embed all the sentences of Wikipedia summary with SBERT, we used 7 GeForce GTX TITAN X GPU and took 1.5 hours to encode all the sentences. Due to the scale of the dataset, we located answers to questions by using the method of approximate nearest neighbour. The index was trained and built using the inverted file with exact post-verification for 4 hours \cite{JDH17}. After building the index, 3 minutes were needed to search the set of answers for all the 8,600 WTC questions with one GPU, an average of 0.02s per question. 

The percentage of questions in different types and topics that can be answered with high confidence are shown in Table \ref{tab:qa-result}. There are 24.69\% of the questions in WTC that can be answered by QBERT.

\begin{table}[]
\caption{Number of questions in WTC can be answered with high confidence over the number of the groups.}
\centering
\resizebox{\textwidth}{!}{%
\begin{tabular}{|l|l|l|l|l|l|l|l|l|l|l|}
\hline
                                 & \textbf{how} & \textbf{what} & \textbf{when} & \textbf{where} & \textbf{which} & \textbf{who} & \textbf{why} & \textbf{if} & \textbf{other} & \textbf{All} \\ \hline
\textbf{Business \& Finance}     & 32/121       & 44/100        & 4/16          & 5/18           & 0/0            & 12/26        & 42/191       & 2/30        & 34/136         & 175/638      \\ \hline
\textbf{Computers \& Internet}   & 6/34         & 2/9           & 0/3           & 0/2            & 0/0            & 0/3          & 1/18         & 0/5         & 4/34           & 13/108       \\ \hline
\textbf{Education \& Reference}  & 42/132       & 36/81         & 2/8           & 1/11           & 2/2            & 14/50        & 14/84        & 1/16        & 14/68          & 126/452      \\ \hline
\textbf{Entertainment \& Music}  & 9/55         & 17/56         & 2/10          & 4/10           & 0/0            & 4/12         & 7/80         & 1/39        & 22/108         & 66/370       \\ \hline
\textbf{Family \& Relationships} & 7/44         & 9/32          & 0/8           & 3/8            & 0/0            & 0/1          & 16/95        & 2/14        & 13/68          & 50/270       \\ \hline
\textbf{Health}                  & 15/159       & 12/66         & 4/18          & 0/10           & 0/0            & 0/6          & 61/299       & 3/34        & 15/84          & 110/676      \\ \hline
\textbf{Politics \& Government}  & 3/23         & 6/18          & 2/7           & 1/2            & 0/0            & 1/5          & 12/57        & 1/22        & 6/51           & 32/185       \\ \hline
\textbf{Science \& Mathematics}  & 416/1355     & 240/646       & 21/88         & 27/99          & 6/15           & 16/58        & 339/1107     & 51/392      & 186/918        & 1302/4678    \\ \hline
\textbf{Society \& Culture}      & 21/142       & 57/159        & 4/23          & 4/21           & 0/0            & 9/52         & 49/286       & 13/108      & 51/237         & 208/1028     \\ \hline
\textbf{Sports}                  & 11/47        & 5/14          & 2/5           & 0/0            & 0/0            & 4/15         & 4/48         & 0/7         & 15/59          & 41/195       \\ \hline
\textbf{All}                     & 562/2112     & 428/1181      & 41/186        & 45/181         & 8/17           & 60/228       & 545/2265     & 74/667      & 360/1763       & 2123/8600    \\ \hline
\end{tabular}}
\label{tab:qa-result}
\end{table}

\section{DISCUSSION OF RESULTS}
“Project What If” was launched in 2017 across Bristol and involved thousands of people. Its aim was to focus on the questions that were most asked by ordinary Bristolians, rather than on the answers, to see what they said about the local Community.

The automated analysis of that corpus, enabled by QBERT in Section 4, revealed that more than half of the questions are in the domain of Science \& Mathematics (54.10\%), followed by Society \& Culture (12.57\%), and then by Health (7.70\%). The most frequently asked type of question is of the type WHY (26.34\%) followed by HOW (24.56\%).

By navigating the IF question, we observed that most of the questions are counterfactual such as “What if we never went to sleep?”, “If you could hear in space, how loud would the Sun be?”. However, the corpus also contains a number of factual questions, as would be expected in a science-centre setting. For example, “I’d like to know if atoms are made up of other atoms.”.

Furthermore, we calculated the $P(type, topic)$ and $P(type)*P(topic)$ to understand the associations between type and topic. The question-type WHO is strongly associated with Education \& Reference and with Sport because the $P(Who, Sport)$ is 3 times larger than the probability of $P(Who)*P(Sport)$. The topic Education \& References strongly associates with types: how, what, when, who. The type IF associates strongly with Politics \& Government, but not with Education \& Reference.

One limitation of pre-training with Yahoo! Answer dataset is that it only takes the top 10 topics from Yahoo! Answer regardless of all other possible topics. During labelling the topic for WTC corpus, we noticed that many questions did not belong to any of the groups in YT. QBERT can be improved with more question topics or labelling the unknown topics.
After applying QBERT, we found answers for 2,123 questions in the WTC corpus. WH questions, such as WHICH, WHAT, HOW, WHERE, and WHO are more likely to be answered by Wikipedia Summary comparing to IF questions and yes/no questions. Due to the QBERT mechanism, the answer is supposed to be one sentence from Wikipedia. In this case, factoid questions which can be answered with fact expressed in a short sentence are more likely to be answered. Furthermore, non-factoid questions, like some of the WHY or IF questions, that require more explanation in the answer are harder to find one sentence answer from a knowledge source. More than 50\% of the Education \& Reference, Science \& Mathematics questions, and Business \& Finance can be answered with confidence by Wikipedia Summary. However, QBERT can only answer around 12\% and 17\% of the questions in Computers \& Internet and Politics \& Government, respectively. 

Here are some examples of the answers giving by QBERT. The questions are from the WTC corpus, and the answers are from the Wikipedia Summary.

\begin{itemize}
    \item \textbf{Q1:} How old is the oldest tree in the world?
\item \textbf{A1:} A scientific investigation in 1965 of the tree’s rings indicated that the tree has an estimated age of 1450-1900 years, and may well be the oldest living oak in northern Europe. (Score: 0.817)
	\item \textbf{Q2:} In the future, will robots gain conciousness?
\item \textbf{A2:} Throughout history, it has been frequently assumed that robots will one day be able to mimic human behavior and manage tasks in a human-like fashion. (Score: 0.775)
\item	\textbf{Q3:} How do birds lay their eggs?
\item \textbf{A3:} They lay their eggs into the wet dangling roots of plants. (Score: 0.788)
\end{itemize}

From the question answering pair found by QBERT, we notice that for questions lacking in attributes such as Q1 in the example, the retrieved answer is adapted to a specific attribute rather than a general situation as human understanding. Another barrier in QBERT is that the answer sometimes is not included in one single sentence. For example, in A3, the original summary is “Zygonyx is a genus of dragonflies … They lay their eggs into the wet dangling roots of plants.”. However, the QBERT is only able to retrieve the most relevant sentence. In this case, QBERT fails to capture the entity, which is more important in the question.

We also observe that some answers with high confidence (over 0.85 cosine similarity) are similar questions that QBERT found in Wikipedia. For example, QBERT is tricked by a sentence in Wikipedia “Why Does the Sun Shine?” and considers it the answer to the question “Why is the sun bright?”.

\section{RELATED WORK}
Question answering is always a challenging research task in NLP. Meanwhile, question pre-processing like topic classification and similar question classification is critical to a large scale question answering system.

The pre-trained language models earn state-of-the-art results in many NLP tasks. The model we used, BERT \cite{devlin2018bert}, and its modified models have leading performance in classification \cite{mccreery2020effective,sun2019fine} and question answering \cite{lee2019latent, guu2020realm, karpukhin2020dense}. 

For question answering, we identified our research as open-domain and open book answer retrieving.  The system was designed to infer the correct answer from knowledge sources like Wikipedia in a concise sentence. Comparing to answering question using a knowledge base, open-domain QA is more challenging in using large-scale knowledge sources and machine comprehension. Previous research \cite{lee2019latent,guu2020realm, karpukhin2020dense, chen2017reading,yang2019end} leveraged a retriever-reader or retriever-generator that retrieved the relevant passage from the knowledge source and extracted an answer span from the passage. The passage can be a document, paragraph, sentence or fixed-length segment. However, this two stages system is computationally expensive. Inspired by DenSPI \cite{seo2019real}, QBERT encodes all the sentences in the knowledge base and searches the most relevant sentence with the query. In addition, we perform approximate nearest neighbour search to reduce the searching time.

Some researchers use Glove word embedding \cite{pennington2014glove} or BERT [CLS] token as sentence embedding to encode the questions and the knowledge base. Instead, we trained S-BERT \cite{reimers2019sentence}, a sentence embedding network that fine-tuned BERT with similar sentences, to retrieve answers. Because S-BERT outperforms Glove and BERT [CLS] in textual similarity tasks. Besides, it reduces the complexity of embedding sentences with BERT.

\section{CONCLUSIONS}
What did the corpus of questions collected by WTC reveal about the Community that generated it? This was the last question that remained unanswered, and we hope that our AI analysis can provide a first insight.

QBERT, a new generalist model for question-content analysis was applied to the WTC corpus. In the results, we see that the contributors to “Project What If” were very interested in Science, Society, and Health; and asked many questions of the WHY and HOW type. This is not surprising within the setting of WTC as a science institution. But the next finding revealed a lot more: questions of the type IF tend to relate to Politics \& Government topics and not with Education \& Reference topics. Are Bristolians exploring alternative ways to be a Community?

Curiosity about Society \& Culture is also very revealing. This is an emerging theme in the sector of science centres, where there is an ongoing discussion about expanding from Science Centres to Science \& Cultural Centres. More generally, there is a movement in the sector currently to explore society and culture alongside traditional science such as Biology, Engineering, Chemistry etc. This seems to be reflected in the kind of questions Bristolians have been asking.

In our modern world, access to information is easy and comprehensible through digital and online channels. Science centres have therefore been challenged to adapt to this changing environment, incorporating social sciences and perspectives from different cultures and presenting a space for exploration of ideas rather than just answers. This is why We The Curious has based “Project What If” on questions, exploration and curiosity, rather than just education and knowledge sharing. These findings support the rationale and aims of these changes. 

With QBERT, more than 50\% of the questions from Science \& Mathematics and Education \& Reference topics have been answered. Moreover, the QA system can answer a high percentage of WH questions except for WHY. QBERT is also computational efficient during retrieval answer from Wikipedia. It takes 0.02s per question. Although QBERT managed to answer 43.8\% of the questions, there are still some limitations with the QA model. We can further explore the QA system in the future to overcome these deficiencies. One of the possible ways is that we can encode the paragraph instead of sentences for question answering.

\textbf{Note:} The anonymised and moderated dataset is available from We The Curious on reasonable request for research purposes. Please contact information@wethecurious.org.

\section*{Acknowledgment}
We acknowledge Anna Starkey for her central role in “Project What If” and being part of the team who initiated this research, and David May for bringing us all together and providing his insight along the way.

\bibliographystyle{unsrt}  
\bibliography{references}  

\end{document}